\documentclass{article}

\usepackage{arxiv}

\usepackage[utf8]{inputenc} 
\usepackage[T1]{fontenc}    
\usepackage{hyperref}       
\usepackage{url}            
\usepackage{booktabs}       
\usepackage{amsfonts}       
\usepackage{nicefrac}       
\usepackage{microtype}      
\usepackage{cleveref}       
\usepackage{lipsum}         
\usepackage{graphicx}
\usepackage{natbib}
\usepackage{doi}

\title{On the Semantics of Large Language Models}
\date{}

\newif\ifuniqueAffiliation

\ifuniqueAffiliation 
\author{{Martin~Sch\"ule}\\
}
\else
\usepackage{authblk}

\author[1,2]{%
	Martin~Sch\"ule\thanks{\texttt{scli@zhaw.ch}}%
}
\affil[1]{Institute of Computational Life Sciences, Zurich University of Applied Sciences, Switzerland}
\affil[2]{Institute of History and Philosophy of Science and Technology (IHPST), Université Paris 1 - Panthéon-Sorbonne}
\fi

\renewcommand{\headeright}{}
\renewcommand{\undertitle}{Preprint. Published in Intellectica, 81, (pp.15-36), 2024.}

\begin{document}
\maketitle

\begin{abstract}
Large Language Models (LLMs) such as ChatGPT demonstrated the
potential to replicate human language abilities through technology, ranging from text
generation to engaging in conversations. However, it remains controversial to what
extent these systems truly understand language. We examine this issue by narrowing
the question down to the semantics of LLMs at the word and sentence level. By
examining the inner workings of LLMs and their generated representation of language
and by drawing on classical semantic theories by Frege and Russell, we get a more
nuanced picture of the potential semantic capabilities of LLMs.
\end{abstract}


\section{Introduction}
The release of ChatGPT has made the potential of technical systems to
replicate human language abilities visible to a large public. Large language
models (LLMs) like ChatGPT can generate text, answer questions, and engage
in conversations, creating an impression of language understanding. However,
debate continues over their true capabilities and comprehension. Do they
understand language in any meaningful way, or are they merely “stochastic
parrots” \citep{bender2021}? The field of AI and with it LLMs will continue
to evolve and due to their potential economic, social, and political impact, as
well as their inherent scientific and philosophical intrigue, one needs to get a
better grasp on the critical question: do LLMs understand language and if so in
what sense? This raises issues about their trustworthiness, ability to reference
truth, and propensity to “hallucinate”; issues which have a certain urgency
giving the growing usage of LLMs. 

Part of the endeavour is to understand how LLMs work and what their
capabilities are. LLMs have been developed in the sub-field of natural language
processing (NLP) and use deep learning (DL) techniques or models. DL models
are in principle complicated neural networks (NN).\footnote{In this article, we will use the terms neural network (NN) and deep learning (DL) model interchangeably
unless specified otherwise.} Now, such models can be
studied in two ways. The first approach treats the system as a kind of black box,
which is studied through its input-output behaviour. There are numerous studies
that follow this approach, subjecting the LLM to a battery of tests to evaluate its
performance \citep{bubeck2023, chang2024}. Basically, the LLM is
treated as a kind of an alien entity which is being interrogated. The prototypical
example is the Turing Test \citep{turing1950}, a question-and-answer game
designed to determine whether responses are generated by a human or a
machine. The second approach tries to investigate the inner workings of a LLM,
the underlying mechanisms, by “opening the box” to eventually draw
conclusions about their capabilities. In this study we follow the second approach
as we are specifically interested in the representations generated by LLMs and
their possible interpretations.

DL models have fundamentally straightforward mathematical descriptions.
Additionally, the trained model and its parameters can, in principle, be
examined, making the inner workings of LLMs accessible for study. However,
due to the complexity and scale of the models, the vast and complex datasets
they may be trained on, and their interactions with the world and users, this is
not a straightforward endeavor. The sub-field of explainable AI aims to analyze
DL models by developing specific techniques for explainability \citep{hassija2024}. Yet, the concepts we are interested in such as “understanding” are seldom
addressed in these technical studies. What we need, therefore, is an approach
that develops conceptual interpretations of LLMs beyond mere technical
descriptions.

In AI, some conceptual vocabulary with terms like “intelligence” and
“semantics” are also used, yet descriptions in these terms can be vague,
potentially even misleading. How can we address the question “What do LLMs
understand?” without considering the definition of understanding itself?
Philosophical literature has grappled with these questions for centuries; ignoring
them seems presumptuous. While philosophy may not offer definitive answers,
it should help clarify and refine these questions. The challenge rather lies in the
vast volume of the accumulated literature on the notion of “understanding” while
at the same the technical development progresses rapidly. We need to find some
middle ground to navigate this scenario effectively.

We address this challenge with a two-fold approach. First, we aim to identify
certain stable elements within LLMs' approach to language and base our analysis
on these elements. Second, we narrow the question of “understanding” to the
semantic capabilities of LLMs at the word and sentence levels. By gaining a
better understanding of how LLMs represent word and sentence meanings, and
how we can interpret these representations, we also progress toward the broader
question of “understanding”. Our analysis will be in terms of what is known as
the “classical semantic theory” which goes back to Frege and Russell and we will later discuss why this is relevant, even though other theories might align
more closely with how LLMs operate. Clearly, there is room for further
investigation in this area.

The article is structured as follows. First, we identify the essential elements
and concepts involved in LLMs’ representation of language. Next, we examine
the semantic theories of Russell and Frege, focusing on the components critical
to our purposes. With this analysis and conceptual framework, we revisit the
questions about LLM semantics, situating them within a relevant technical and
philosophical context.

\section{What are LLMs?}
\label{sec:headings}

We first need to look at the inner workings of LLMs. Without this, we cannot
make a precise case regarding the semantic qualities of LLMs. Our explanations
will be general and focus on aspects essential for further discussion. Readers
familiar with these topics may skip to the summary.

As mentioned, discussing LLMs involves navigating a rapidly evolving
research field. What may be considered state-of-the-art today may soon become
outdated. However, we believe that certain stable elements in the LLM approach
to language can be identified:

\begin{enumerate}
\item The probabilistic approach to modeling language.
\item The distributed representation of language.
\item The use of connectionist architectures such as neural networks (NN).
\item The use of large-scale models.
\end{enumerate}

In LLMs, all these elements are present and interwoven; however, it is
beneficial to consider them separately for the time being. Also note that
historically these elements correspond to different eras of language modeling in
NLP and AI.

Other aspects like specific model architectures will likely change or are
currently undergoing changes, so it is essential to be aware that things can evolve
quickly. Nonetheless, it is at times unavoidable to explain certain things in more
detail such as the exact nature of the training or some specific model architecture
to gain an understanding of the inner workings of current LLMs even though
these aspects may change.

\subsection{Language Models}
Let’s discuss the probabilistic approach first, which is closely linked to the
concept of a language model (LM). An LM is a probabilistic representation of
language, which goes back to Shannon \citep{shannon1951}, and has been used in
natural language processing (NLP) for decades \citep{jurafsky2024}.
Before we discuss LMs further let’s briefly look at the pre-processing of text
we face in NLP tasks, restricting ourselves to the case of written text. First of
all, text must be converted into a suitable form to be processed. This process is
called tokenization: text units are assigned so-called tokens. For simplicity, we
will assume that these tokens represent words, although they could also represent
sentences, and in practice, more efficient sub-word tokenization schemes are
used \citep{radford2019}. Given a text, it is broken down into word tokens; a sentence then becomes a sequence of tokens, for example, the sentence “the cat
sat on the mat” is tokenized to [“the”, “cat”, “sat”, “on”, “the”, “mat”].

A language model associates such sequences with a probability distribution,
$P(x_1, x_2,..., x_T)$ which indicates the probability of the sequence of words or
tokens.

How can this distribution be calculated? According to probability theory
\begin{equation}
P(x_1, x_2, ..., x_T) = \prod_{t=1}^{T} P(x_t | x_{t-1}, x_{t-2}, ..., x_2, x_1).
\end{equation}

The distribution can thus be built up successively. One starts with the probability
for a token, then calculates the conditional probabilities for the next tokens, and
so on. This naturally leads to an autoregressive model: based on the previous
words, the next word is generated, then the subsequent word, and so forth, given
the corresponding conditional probabilities and a sampling technique which
samples the actual tokens from the distributions.

Such language models are very useful and have been used in NLP long before
the advent of LLMs for tasks like speech recognition \citep{jelinek1990}, etc. Also
a LLM is essentially a language model.

How can these probability distributions be calculated? Before neural
networks (NN) were used for this purpose, purely statistical models, known as
N-gram models, were employed. N-gram models are essentially Markov models
that estimate words from a fixed window of previous words, called the context,
where probabilities are computed from a given text corpus in a frequentist
manner \citep{jurafsky2024}. This statistical approach has certain
disadvantages, notably the curse of dimensionality, which makes modeling joint
distributions challenging. For instance, modeling the joint distribution of 10
consecutive words with a vocabulary of 100,000 words leads to an enormous
number of potentially 1050 free parameters \citep{bengio2000}. The n-gram
approach thus typically restricts the context size to say 3 words, missing out on
a lot of context. In contrast, the NN approach learns the probability function for
word sequences during training and can generate distributions for sequences not
seen during training.

Language models, as autoregressive models, can readily be used for
generative purposes. The model generates probabilities for the next tokens,
sampling a token to become part of the context, and then proceeding to generate
the next token, and so on. This has been criticized because errors in
autoregressive predictions tend to accumulate which might be one of the reasons
LLMs can venture into “hallucinations” \citep{lecun2023}. In general terms
however, it seems unlikely that the probabilistic modeling of language will be
abandoned, as language usage evidently has a stochastic component. When I
begin a sentence, there are various possibilities how to continue it, based on the
previous words, and this seems to be an essential characteristic of language. In
NLP, the most effective way to model such features is through a probabilistic
approach.

\subsection{Distributed Representations}
A distributed representation, also called an embedding, means representing
words or tokens as vector space representations. Tokens are no longer
represented as symbols but as feature vectors with, in principle, continuous
values, mathematically expressed as $x \in \mathbb{R}^d$, where an embedding dimension $d$
needs to be chosen, which defines the number of features. A token is thus
represented as a point in a high-dimensional space, the embedding space, also
called latent space in this context. The joint probability function of word
sequences, as used in an LM, is then expressed in the form of the feature vectors
of these word sequences, and a model simultaneously learns the word feature
vectors and the parameters of this function.

The idea of distributed representation has been around in linguistics as
distributional hypothesis: words that occur in similar contexts tend to have
similar meanings \citep{harris1954} and in more technical terms in information
retrieval systems \citep{schutze1992}. In a NN context the concept seems to have
been introduced by Hinton \citep{hinton1986} and for language models in the
seminal paper by Bengio et al. \citep{bengio2000}. Distributed representation
projects language into a vector space, moving away from traditional symbolic or
discrete representations. The approach has several advantages: compact
information encoding, the ability to compare representations through similarity
and distance measures and the potential to overcome the curse of dimensionality.
It can be argued that by assigning similar feature vectors to related words, we
increase the likelihood of a combinatorial number of related sentences which
helps alleviate the curse of dimensionality problem \citep{bengio2000}.
Furthermore, models with embeddings tend to significantly outperform models
without embeddings in NLP tasks.

The distributed representation is inherently linked to the NN approach; it is
essentially a NN-generated representation. That said, it can be considered an
independent idea. In principle, a distributed representation can be generated
independently of a NN, for example, with a statistical approach \citep{pennington2014} or with a shallow NN \citep{mikolov2013}, which does not have to
correspond to the NN in which the embedding is being used.

Today, distributed representations or embeddings are considered essential
components of NN or DL model architectures in NLP. An embedding is required
in the input layers of a DL model to project the discrete token structure onto the
vector space. Typically, the layers preceding the output layers are considered the
embedding or latent space. In principle, however, each layer of a DL model can
be seen as a distributed representation – this would be the horizontal
representation. There is also a vertical component; studies on BERT models,
similar models to autoregressive LLMs, show that information is not encoded in
separate layers, although more abstract concepts seem to be represented in later
layers, but spread across the entire model \citep{rogers2021}.

Given their mathematical advantages, the fact that models utilizing
embeddings empirically perform better on NLP tasks, and their integration with
DL models, it is unlikely that this method of language encoding will be
abandoned in the near or mid-term future. DL models, as discussed in the next
section, are distributive systems and naturally operate with distributed representations and vice versa. Language, possibly reasoning, may operate
symbolically, but their underlying representation and processing appears to be
distributive, at least in the NLP/AI approach to language.

\subsection{Neural Networks}
Neural networks (NN) are loosely modeled after biological neural networks.
According to the connectionist paradigm, it requires many, possibly a very large
number of simple building blocks to collectively achieve complex behavior and
achieve complex cognitive abilities. Currently, state-of-the-art models are
Transformer-based, so we’ll briefly look at these in more detail. In the future,
other models or architectures may replace them.

\begin{figure}[h]
    \centering
    \includegraphics[width=0.5\textwidth]{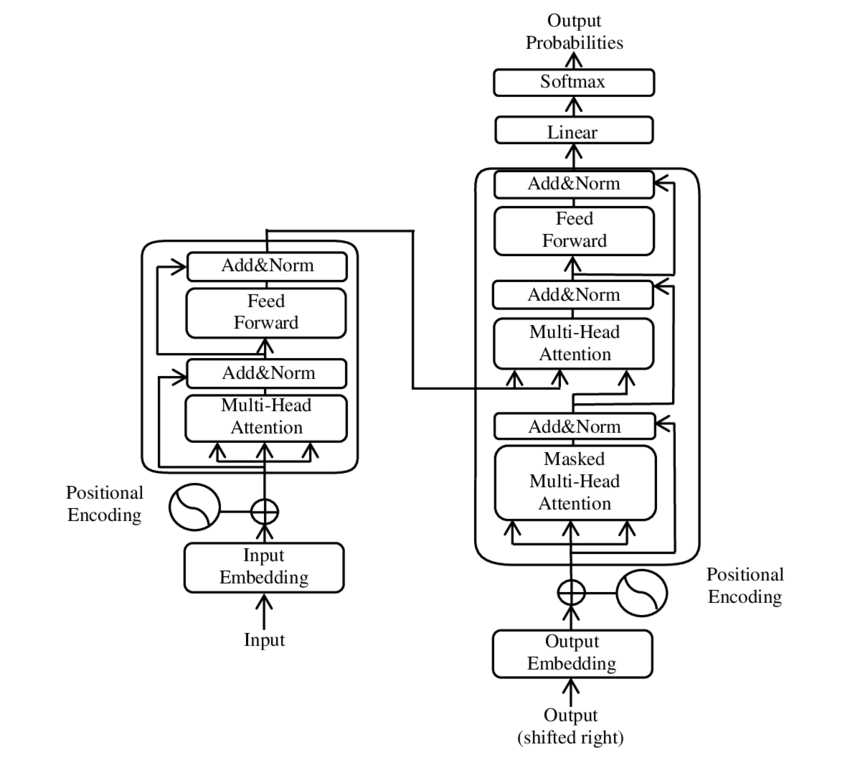}
    \caption{The original Transformer architecture with the encoder (left) - decoder (right) structure. \cite{vaswani2017}.}
    \label{fig:Transformer_architecture}
\end{figure}

Transformers \citep{vaswani2017} are DL models that have set new
standards in NLP tasks. Initially designed for translation tasks, generative
models either use the decoder or encoder part of the original Transformer model
(see Figure \ref{fig:Transformer_architecture}). The details of the model architecture cannot be explained here,
see \citep{vaswani2017}. Two points are noteworthy: First, tokenized text is
mapped to an embedding, forming the input. The model then transforms this
representation while maintaining the format, using it in the final step to perform
a specific task. Second, the key mechanism in transforming the representation is
the so-called attention mechanism, which allows each word in an input sequence
to be compared with every other word in the sequence, thus considering the
context of each word. Unfortunately, this entails significant computational
effort, as compute increases quadratically with the input sequence length.

Transformer-based models generate a context-sensitive representation of
language, meaning that a word will be represented differently depending on its
context. For example, “play” in “the children play” is represented differently
than “play” in “theatrical play”. This representation depends on the context size,
that is, how many words the model is conditioned on, as discussed in regard to
language models. Early models had a context size like 512 tokens \citep{radford2019}, while current models have a context size of several hundred thousand
tokens \citep{achiam2023}. The context length determines and fundamentally
limits how much content is considered when generating a response to an input.

DL models learn language through tasks. A simple task used in training is
the next-word prediction task. Given a text, i.e., a word sequence, the next word
is masked, and the model must predict it. This task aligns naturally with the
structure of a language model, as discussed. The task is particularly suited for
self-supervised learning because it allows for large training datasets where
words are automatically masked. The ultimate goal is to perform well on test
datasets and in inference mode, not only for next-word prediction but also for
further downstream tasks.

NN and DL models operate in a distributive way according to the
connectionist paradigm, where memory and processing occur simultaneously
across many simple interacting nodes or (artificial) neurons. The approach
naturally allows for parallelization. The size of an NN too seems to matter
because it allows for the formation of subroutines and submodules \citep{lepori2023}. We will revisit this point in the next section.

As mentioned, other models might eventually replace Transformer-based
models, potentially incorporating some form of recurrent architecture \citep{gu2023}. This remains to be seen. However, it is expected that various types
of neural networks, i.e., connectionist architectures in the broader sense, will
continue to be used. Their flexibility, parallelization capabilities, and
performance is unmatched. Another reason is that the concept of distributed
representation is essential for the DL approach to language. Distributed
representations are made for DL models, and vice versa. In principle, the two
could be separated, as mentioned, but it seems inefficient to use distributed
representations without connectionist models.

\subsection{Large Language Models}
Large language models (LLMs) are advanced language models that use
neural networks (NN) and are scaled in regard to the amount of data, model size,
and computational resources. Currently, Transformer-based architectures
dominate the field and are typically trained on the next-word token prediction
task.
In the next-word token prediction task, the model is given a sequence of word
tokens, and the task is to predict the next word token. Here are some examples:\\
$[$'the', 'cat', 'sat', 'on', 'the'$]$ $\rightarrow$ $[$'mat'$]$\\
$[$'what', 'is', 'the', 'capital', 'of', France, ?$]$ $\rightarrow$ $[$'Paris'$]$\\\
$[$'translate', 'to', 'French', 'sea', 'otter'$]$ $\rightarrow$ $[$'loutre de mer'$]$

Next-token prediction, while seemingly a basic task, seems to offer an
optimal level of complexity for effectively scaling language models. Also, it does not mean that the model is limited to this specific task. As hinted in the
examples, the task can include additional tasks; the model learns further tasks
beyond the primary task. This is known as in-context learning. For example, the
model learns a translation task in the above example, even though the primary
task is next-token prediction.

In the inference mode, i.e., when tokens or text are generated, the model takes
an input text (the prompt) and continues it in an autoregressive manner. This
means a token is generated, based on probability distributions and sampling
techniques, which is then added to the input. Based on this extended input, the
next token is generated, and so on, as it happens in autoregressive modeling with
LMs.

Commercially used LLMs, like ChatGPT \citep{openai2022}, Claude \citep{anthropic2022}, or Mistral \citep{mistral2023}, typically incorporate an additional mechanism known as alignment, which aims to adjust the output according to certain guidelines. Various techniques exist which may
involve human feedback such as the RLHF algorithm \citep{christiano2017, rafailov2024}. However, since this secondary adjustment of LLM outputs
does not alter significantly the internal representation but rather aims to control
the output, we believe the core semantic capabilities of LLMs remain largely
unaffected by current alignment techniques. This perspective is supported by
experiments conducted with earlier GPT models, which demonstrated similar
semantic functionality, before the implementation of alignment methods.\footnote{For a contrary view see Mollo \& Millière \citep{mollo2023}.}

Studies have shown that Transformer-based LLMs scale with size, meaning
their quality improves with the size of the dataset, the model size, and the
computational resources used \citep{kaplan2020}. This likely applies to other
types of models as well, but remains to be seen. However, a certain size seems
necessary for several reasons. Firstly, storing information obviously requires
memory space, i.e., in the case of DL models, a certain number of neurons and
connections are needed. In-context learning also seems to demand an adequate
size of both the training data and the model itself to facilitate the development
of subroutines and submodules \citep{lepori2023}. Finally, while the
connection is loose, it seems that the cognitive abilities of animals also scale,
more or less, with brain size \citep{reader2002}. In a connectionist
paradigm, which uses simple computational elements, it seems apparent that
networks using such components must reach a certain critical size to function.

\subsection{Summary}
Large language models (LLMs) represent a sophisticated approach to
language modeling, based on probabilistic methods, distributed representations,
neural networks (NN) or deep learning (DL) models, and large-scale
frameworks. We have argued that while technical details continuously change
and evolve, these four elements form a stable core to the DL or AI approach to
language.

At the core of LLMs is a probabilistic approach to language. Specifically,
language models (LM) calculate word sequence probabilities successively, leading to autoregressive models, using Transformer-based DL models. While
the details of the approach may change, its probabilistic nature is likely to remain
predominant.

The training of LLMs is task-based and data-driven. Typically, models are
trained on next-word token prediction, which offers several advantages. In
principle, models could be trained on alternative tasks, as demonstrated by
BERT models \citep{devlin2018}, likely leading to representations with similar
properties. In this sense, the resulting representation is both task-driven and task-agnostic. LLMs generate a representation of language that can be used for further
tasks, such as text generation, but also numerous other NLP tasks. The quality
of a model is assessed by how well it can perform across these tasks.

LLMs generate distributed, high-dimensional vector space representations,
also known as latent space. In these spaces, tokens, such as word tokens, are
represented as points. Now, a point, a single vector representation, is not
inherently meaningful; it is just a bunch of numbers. Only in relation to other
points in that space we gain a meaningful and context-sensitive representation.
The representation is context-sensitive in two ways. First, in a narrow sense as
employed in the LLM literature, the specific vector representation of a word
depends on the context in which it appears. Second, in a broader sense, the
representations relate to the entire model and dataset the model is trained on,
incorporating the information in the dataset throughout the model as induced by
the training task.

The representations generated by LLMs depend on the datasets with which a
model is trained, including bias and misrepresentations \citep{bender2021}.
However, when datasets and models are sufficiently large, models will likely
generate similar representations. For example, it will always give the same
answer to the question “What is the capital of France?”, unless there is a specific
answer elicited such as an attempt at a joke. This is also underscored by the
observation that different LLMs may vary in details but generally perform
similarly in broad terms. The representation is also independent of human
intervention: different prompts may evoke different answers from the model, yet
the underlying representation capacity of the trained model is unaffected, unless
the model is actually trained or fine-tuned on human feedback.

The generated representation can be considered objective and transparent in
the following sense: Given similar datasets and training methods, the system will
generate a representation similar to other models, meaning the representation is
somewhat independent of the specific model used as well as independent of the
prompting, as mentioned above. The trained model is also transparent in the
sense that the trained model with its parameters can be inspected by anyone,
provided they have access. In this sense, there is no privacy to the model.

LLMs are scaled language models that leverage large datasets and
computational resources. Usually trained through the next-word token prediction
task, they generate representations allowing for in-context learning and the
development of additional capabilities. In this sense, they exhibit emergent
capabilities, akin to what is referred to as weak emergence in complex systems
\citep{bedau1997}. The performance, at least of Transformer-based LLMs, increases
with the dataset size, the model size and corresponding computational resources.
It appears that this is a general inherent property of connectionist models that
have the power to exhibit new, emergent properties.

\section{Semantic Theories}
Having reached a solid understanding of how language is represented in large
language models (LLMs), we can now turn to their semantic properties. As
motivated in the introduction, we want to take recourse to philosophical theories
of meaning to get a thorough understanding of the semantic capabilities of
LLMs.

Ultimately, we want to advance on the question of “What do LLMs
understand?”. But the concept of understanding is broad and diverse and touches
on general epistemological theories. As a first step it seems more useful to
address a seemingly simpler question: when do we understand the meaning of a
word or sentence? However, since we want to separate the grasp of meaning
from meaning itself, as various semantic theories do, we consider the yet more
fundamental question: what constitutes the meaning of a word or sentence?
Specifically, in the context of LLMs, we thus look at the question: what is the
meaning of a word and sentence, as represented by LLMs. We examine this
question through the lenses of the semantic theories by Russell and Frege, which
we will motivate further below. As we shall see, following this more precise
semantic theories actually allows to return to the question of understanding in a
more precise but also narrower sense.

Now, as alluded to in the introduction we can study deep learning (DL)
models such as LLMs either by external criteria or by internal criteria. This also
applies to the semantic properties of LLMs. In AI, if an LLM passes semantic
tests, it is regarded as possessing semantic capabilities. For instance, a natural
language processing (NLP) task like sentiment analysis, which aims to
determine whether a text conveys positive or negative sentiment, is considered
to have a semantic dimension. Consequently, a model capable of performing
such tasks is viewed in the fields of NLP and AI as having semantic capabilities.
This is an example of the typical understanding of the term semantics in AI
research and applications.

We want to go beyond this broad notion of semantics by incorporating the
internal representations of LLMs in the discussion and by using a more refined
terminology from the philosophical literature by drawing on philosophical
theories of meaning. Now, a challenge we face is not only the rapidly evolving
field of AI, NLP, and LLMs, but also the diverse array of theories of meaning.
We have to find some middle ground and establish a starting point for our
exploration which we find in the so-called classical semantic theory.

Classical semantic theory goes back to Frege and Russell and emerged from
efforts to resolve fundamental problems in mathematics and logic but came to
influence philosophy of language and, in particular, theories of meaning
significantly \citep{frege1879, frege1884, russell1903, haack1978}. There are
two principal reasons why it is worthwhile to discuss these theories. First, the
theory is very much at the center of 21st philosophy’s take on the notion of
meaning. Most subsequent work either builds upon, discusses or critiques the
theory \citep{burge1990}. Wittgenstein’s pragmatic turn on language and meaning
\citep{wittgenstein1958}, for instance, can be seen as a reaction to Russell and Frege as well as his own earlier work on language and meaning \citep{conant1998}. It
seems reasonable to study the core tenets of modern semantic theories first,
before specific subsequent theories, even though the latter might align more
closely with how LLMs function. Second, and more importantly, the classical
theory particularly focuses on how linguistic expressions contribute to the
meaning and truth or falsehood of sentences, a specific but fundamental aspect
of language. Now, it is precisely this aspect of language and meaning LLMs
struggle with; they do not struggle with fluency, conversations and engagement,
but rather with their tendency to “hallucinate”, i.e., they struggle with getting
facts, logic and reasoning right. Thus, classical semantic theory may offer
valuable insights into addressing these issues.

When it comes to the discussion of the classical semantic theory, there is a
lengthy and convoluted history of diverging interpretations; we must inevitably
restrict ourselves to certain essential key points of these theories which are
fundamental to our analysis. The aim is to give a concise but correct rendering
of the theories, to the extent where this is possible, and indicate where we set on
a maybe more particular point of view. We first discuss the general terms
involved and then more specifically Russell’s and Frege’s views.

\subsection{Reference and Meaning}
A fundamental notion of semantic theories is the notion of reference. The
reference assigns meaning to linguistic expressions in virtue of denoting the
objects or entities the expressions refer to. For example, in the sentence “The
evening star is a planet.”, the name “evening star” refers to a specific celestial
body; this reference provides the expression “evening star” with its meaning.
Usually, it is assumed that there is a single referent associated with an
expression. Reference or denotation also explains how the meaning of an
expression affects the truth value of sentences containing that expression. The
difference in truth value between “The evening star is a planet.” and “The North
Star is a planet.” arises from the different references of the names involved.
Simply put, reference “grounds” linguistic expressions in reality.

A theory of meaning which relies solely on the notion of reference however
faces several problems. Many words and sentences are ambiguous or vague and
refer to different things in various contexts, complicating the notion that each
word has a single referent. The theory also cannot easily explain why two terms
with the same referent, such as Frege’s example of “morning star” and “evening
star”, can differ in meaning or convey different information, although they refer
to the same object (the planet Venus). Also, it struggles to account for
meaningful sentences in which phrases refer to non-existent objects, such as in
Russell’s example “The current king of France is bald.” \citep{russell1905}. Many
names or terms, like fictional characters or hypothetical objects, etc. lack real-world referents but are still meaningful in language. Due to these issues, the
referential theory of meaning is seen as insufficient to fully explain how
language and meaning function.

Frege addresses this issue by introducing the now classical distinction
between meaning as sense and meaning as reference. The sense of an expression
is the kind of meaning that refers to “the mode of presentation” or the way in which a referent is indicated, while reference is the actual object or truth value
to which the expression corresponds \citep{frege1892a, frege1948}. This distinction
explains how the sentence “The evening star is the morning star.” has meaning
and epistemic value although from a referential viewpoint it is a tautology.
However, the notion of sense introduces new, debatable problems; we return to
this below.

The core of classical semantic theory deals with propositions. Propositions
are denoted by declarative sentences which have subject-predicate form and can
possess a truth value. In the declarative sentence “The evening star is a planet”,
“the evening star” is the proper name and “is a planet” is the predicate. Although
declarative sentences constitute only a small part of language, they are an
essential part of it as they allow for factual and truthful assertions. This analysis
largely focuses on these sentences, as they constitute a specific but important
aspect of language and are something that LLMs should handle accurately.

Let us now discuss the classical semantic theory in more detail. We begin by
reviewing Russell's view, even though Frege's work preceded and influenced
Russell, because Russell's theory is somewhat conceptually simpler, albeit not
in technical terms. Then, we discuss Frege’s theory in more detail.

\subsubsection{Russell}
Russell’s views on the theory of meaning are primarily detailed in his works
“Principles of Mathematics” \citep{russell1903} and “On Denoting” \citep{russell1905},
but they are also dispersed across other works and have evolved over time. It is
not easy to pinpoint his final conclusions, especially since they are also debated
in secondary literature. Here, we must focus on some fundamental aspects of his
theory.

Russell also distinguishes between what he terms denotation and meaning,
corresponding with the notions of reference and sense as previously introduced.
For example, in “Principles of Mathematics” \citep{russell1903}, he asserts that the
term “the present King of France” has meaning even though it has no denotation,
if there is no present King of France. In order to understand the meaning of a
sentence, Russell then argues that its logical form must be uncovered through
logical analysis.

Russell elaborates this in “On Denoting” by introducing the theory of
descriptions. The theory offers a way to handle sentences referring to non-existent or ambiguous entities. For instance, the denoting phrase “the present
King of France” can be meaningfully analyzed even if there is no actual King of
France. He argues that denoting phrases (e.g., “a man”, “every man”, “the man”)
must be contextually understood within a sentence, deriving their meaning from
the logical form of the sentence in which they appear. The phrase “the present
King of France” has no meaning on its own (contrary to the theory in “Principles
of Mathematics”), only as part of a sentence it gains meaning. The logical
analysis then reveals that the phrase “the present King of France,” rather than
referring to an individual, serves as a kind of placeholder that introduces a
quantificational framework into the sentence. In the case of the statement “The
present King of France is bald”. We actually have a conjunction of statements;
if one of the statements turns out to be false (namely that here is an existing individual who is currently the King of France), the whole statement is false.\footnote{In the case of the statement “the present King of France is bald,” we have the conjunction of
statements: 1. that there exists an entity x such that x is the current King of France; 2. that for all entities y,
if y is the current King of France, then y must be identical to x, i.e., there is at most one entity which is
currently King of France; and 3. that for every x that is currently King of France, x is bald. The analysis
reveals the statement to be false, as there is no existing individual who is currently the King of France.}
Although the phrase “the present King of France” has no meaning on its own,
the sentence “The present King of France is bald.” has a meaning and is
ultimately false. In Frege’s theory, both the phrase and the sentence would have
meaning, but the meaning of the sentence would be indeterminate unless we
presuppose that the phrase “the present King of France” has a reference \citep{frege1892a, frege1948}.

In this context, Russell also differentiates between knowledge by
acquaintance and knowledge by description \citep{russell1905}. In perception, we
can have direct knowledge of perceptual objects, but for most other things, such
as expressed in the statement “the center of mass of the solar system” we rely on
denoting expressions. In his logical framework, the objects of acquaintance are
called logically proper names, directly linked to sense-data, and expressed
through terms like “this” or “that chair”. All other terms must be understood
through logical analysis as exercised above.

The semantic theory also spills over into Russell’s view of how the world is
organized. In his so-called theory of logical atomism \citep{russell1918}, it is
claimed that the world can be understood in terms of simple, indivisible
components (atoms) and the logical relationships between them. According to
Russell, sentences are complexes that express facts about the world, and their
meaning is determined by the components of these facts. The structure of a
sentence should mirror the structure of the fact it represents, which means
understanding a sentence involves recognizing the relationship between its
components. Again, he emphasizes that the meaning of an expression is
determined by its context within a sentence or a larger logical structure.

Ultimately, Russell’s theory of meaning, with its emphasis on logical form,
sentence structure, and reference, is closely tied to his theory of truth. According
to this theory \citep{russell1912}, beliefs are true if a corresponding fact exists and
false if it does not. Both a belief and a fact are considered “complex unities”; if
they share the same order, the same relational structure, they correspond to each
other, making the belief true. Again, to understand the meaning of a sentence
that expresses a belief, Russell argues that one must analyze its logical form and
the relationships between its components within a sentence. This structure is
essential for determining truth, as it allows for a systematic evaluation of
statements based on their correspondence to the facts of the world.

\subsubsection{Frege}
In Frege’s theory of meaning \citep{frege1891, frege1892a, frege1892b, frege1918}, proper names have both a reference (Bedeutung) and a sense (Sinn). A
proper name is a sign or a set of signs that refer to an object. The reference is the
object itself to which the sign refers. The name “evening star”, for example,
designates the planet Venus, thus the reference is the actual celestial body.
Besides the reference, a sign or proper name also has a sense. According to Frege, the sense of a proper name is “the mode of presentation” \citep{frege1892a, frege1948}. The sense is associated with an expression or sign, while the
reference is associated with an object. The proper name expresses a sense and
denotes a reference. The distinction between reference and sense allows Frege
to explain how identity statements like “a=b” can be informative. The
expressions “evening star” and “morning star” are identical insofar as they refer
to the same object but have different senses. The sense of a word can be grasped
by everybody “who is sufficiently familiar with the language or totality of
designations to which it belongs” \citep{frege1892a, frege1948}. A proper name
can have a sense but no reference, however, in regular language use, a reference
is usually assumed \citep{frege1892a, frege1948}.

In addition to the sense and reference of a proper name, Frege also introduces
the notion of conception. The conception is a kind of “internal image” that is
variable and subjective, differing among individuals who hold the conception
\citep{frege1892a, frege1948}. Frege gives the following example: Observing the
moon through a telescope, the referent is the moon itself; the projected image by
the object glass is the sense; and the observer's retinal image the conception. The
conception is subjective, whereas the sense is objective but not the object itself.

A sign, such as a word, can also refer to another sign or word. Then, the sign
is treated as an object itself. This is referred to as oblique or indirect speech. The
indirect reference of a word is its usual sense. For example, in indirect meaning,
the term “morning star” does not refer to the planet Venus; instead, it literally
refers to what is visible in the morning sky.

A sentence, more precisely a propositional sentence, also has a sense. Frege
calls the sense of a sentence the thought (Gedanke) \citep{frege1892a, frege1918, frege1948}. The sense of a sentence is composed of the senses of its parts; thus,
it can be said that anyone understands the thought who “has sufficient knowledge
of the language”. The reference of a sentence, however, is its truth value, i.e.,
the condition that determines whether a sentence is true or false. Now, true
knowledge comes with the thought, the sense of the sentence, and its truth value,
the reference of the sentence.\footnote{``We can never be concerned only with the referent of a sentence; but again the mere thought alone yields
no knowledge, but only the thought together with its referent, i.e., its truth value.'' \citep[p. 217]{frege1948}.} By advancing from the thought to the truth value,
we gain knowledge, a process termed “judgment” by Frege.

Reference is fundamental in Frege’s theory as well. At first glance, it seems
clear what the reference of a proper name is; it is the object to which a name
refers. The difficulties begin when we have names like “the number 2” or
fictional names like “Odysseus”. Consider Frege’s example, “Odysseus was set
ashore at Ithaca while sound asleep.” Despite the uncertainty about the name
Odysseus in this sentence, the sentence still has a sense. However, by being
uncertain about the reference of the name Odysseus, it is questionable whether
the entire sentence has a reference. According to Frege, anyone who considers
this sentence true or false assigns both a sense and a reference to the name
Odysseus. In practice, we often presuppose a reference even if it is nonexistent
or cannot be established.

Determining the sense is also challenging. Is not entirely clear what Frege
means by it since there is no precise definition of sense, only descriptions. Hence, there is considerable debate in the philosophical literature. Some authors
doubt that a sense without reference is meaningful \citep{dummett1981, evans1986}. Dummett describes the sense as “a way to determine its reference” or
“some means by which a reference is determined” \citep{dummett1981}. Some
authors argue that a sense without reference is simply impossible \citep{evans1986}.
However, this does not seem to reflect how language works. It may be that in a
logically pure language, as envisioned in Frege’s logicist programme, it should
be impossible to form a proper name that lacks a reference. But this requirement
does not necessarily apply in ordinary language as Frege acknowledges too
\citep{frege1892a, frege1948}.

It is beyond the scope of this paper to delve into the entire debate. Actually,
we may rather add something to the discussion. For now and for
our purposes, we adhere closely to Frege’s own statements. Let us for the
moment simply list the properties that, according to Frege \citep{frege1892a, frege1948}, the sense possesses:\\

\begin{enumerate}
\item the sense is ``the mode of presentation'' and can be grasped by anyone ``who is sufficiently familiar with the language or totality of designations to which it belongs''  \citep[p. 210]{frege1948},
\item every grammatically well-formed expression representing a proper name has a sense,
\item a word may have a sense but no reference,\footnote{``The words ``the celestial body most distant
from the earth'' have a sense, but it is very doubtful if they also have
a referent. The expression ``the least rapidly convergent series'' has
a sense; but it is known to have no referent, since for every given
convergent series, another convergent, but less rapidly convergent,
series can be found. In grasping a sense, one is not certainly assured of
a referent.'' \citep[p. 211]{frege1948}.}
\item a word with a reference can have several senses,
\item the sense of a word is the common property of many,\footnote{``This constitutes an essential distinction between the conception and the sign's
sense, which may be the common property of many and therefore is
not a part or a mode of the individual mind.''  \citep[p. 212]{frege1948}}
\item in one and the same context, one and the same word should always have the same sense,
\item a sentence has a sense (thought) and possibly, but not necessarily, a truth value,\footnote{``Is it possible that a sentence as a whole has only a sense, but no
referent? At any rate, one might expect that such sentences occur,
just as there are parts of sentences having sense but no referent.'' and ``By the truth value of a sentence I understand the circumstance that it is true or false.''  \citep[p. 215/16]{frege1948}}
\item the sense (reference) of the parts of a sentence is relevant for the sense (reference) of the sentence.\footnote{``If it were a question only of the sense of the
sentence, the thought, it would be unnecessary to bother with the referent of a part of the sentence; only the sense, not the referent, of the part is relevant to the sense of the whole sentence.'' and ``We have seen that the referent of a sentence may always be sought,
whenever the referents of its components are involved; and that this
is the case when and only when we are inquiring after the truth value.'' \citep[p. 215/16]{frege1948}}
\end{enumerate}

Thus, to summarize, the sense of a word is related to linguistic competence,
may or may not have a reference, shows a certain degree of objectivity, and is context-dependent. The sense of a sentence also depends on its constituent
senses and also may or may not have a reference. How this theory fares with the
semantics of LLMs will be seen in the next section.

\section{Discussion}
We have now reached the point where we can compare the representational
and potential semantic capabilities of large language models (LLMs) with the
main strands of Russell’s and Frege’s theories of meaning.

Language, specifically a sentence, is not broken down into its logical
structure by an LLM. Also, neither the individual sentence parts, nor the
sentence as a whole, nor any non-existent logical form as a representation in an
LLM, refer to an external reality. However, in Russell’s view, consistent with
his theory of meaning, realism and referential theory of truth, language should
sensibly mirror an external reality. Such a correspondence does not occur in
purely text-based LLMs, at least not directly. There is no knowledge by
acquaintance as LLMs lack perception, so an LLM does not know what is meant
by the expression “this chair” or “that chair”. Knowledge by description on the
other hand would require logical analysis and comparison with the facts of the
world, which does not happen either.

The notion of reference allows linguistic expressions and their
representations to be “grounded” in reality, but how this happens is debated; in
cognitive science, the problem is known as the “symbol grounding problem”
\citep{harnad1990}. Text-based LLMs, quite obviously, lack reference; based on this
it has been concluded that these systems cannot possess or learn meaning \citep{bender2020}. There are two possible responses to this: First,
reference may not be the only kind of meaning involved here \citep{piantadosi2022}; we return to this point later. Second, it may be the case that there is a kind
of indirect referencing going on which potentially ground LLMs nevertheless,
e.g. by a causal-information mechanism or historical relations \citep{mollo2023}. Giving an extensive review of possible ways of referential
grounding is beyond the scope of this paper; we may however add, in the context
of our discussion, the following observations.

One way to salvage the referential capabilities of LLMs is by directly
incorporating references or allowing textual references. If LLMs can memorize
text passages verbatim, which is possible, and if they incorporate, for instance,
Wikipedia text that correctly refers to the external world, the referential ability
is implicitly incorporated. However, this seems unsatisfactory. We want LLMs
to genuinely generate their own text, not simply memorize existing text. The
second option would allow the internal representation of a complex expression
refer to another complex expression, as in Frege’s indirect referencing scheme.
This does not seem to align with what Russell meant, even if some semiotic
theories postulate that this is actually how referencing works \citep{eco1979}.
Moreover, such a reference would also need to be verified in some way, likely
through another reference. But then we face the question of whether there is
some initial reference or grounding or whether we are lead to a kind of infinite
regress, an infinite semiosis \citep{eco1979, peirce1867}.

An alternative to addressing the limitations of existing text-based models is
the development of multimodal models, i.e., DL models that integrate various modalities such as text, vision, and potentially other modalities via sensory data.
Large multimodal models (LMMs) could then ground their linguistic and
semantic representations in non-textual representations like corresponding
images or sensor data, akin to what has been termed sensorimotor grounding
\citep{harnad1990}. Note however, that such models would still not have direct
access to the world but a mediated access: an access that is fundamentally task-
driven and representational moreover. Also, as we will argue, the issue is rather
that we need to ground sentences, rather than specific representations, because
it is sentences that may refer to truth. But attaining the truth is not an easy task;
ultimately, future LMMs will face the same difficulties as we do in determining
truth.

Other aspects, however, align well with the notion of LLMs having semantic
capabilities from a Russellian perspective. First, it is not excluded that the
representation of LLMs can be transposed into a logical form or at least be used
to construct logically coherent statements. For example, it is possible to modify
the representation to ensure that statements adhere to the law of non-contradiction \citep{burns2022}, which may be a step towards embedding
logical analysis within LLMs. Second, and more generally, Russell’s contextual
principle appears to also apply to LLMs, though not in the logical form per
Russell. A word gains meaning only in the context of other words; more
generally, words and sentences represent in relation to the overall context of the
entire training data and model training. We will return to this point below.

Frege’s theory of meaning is closer to how LLMs work because he
emphasizes the importance of the kind of meaning called sense. The latent space,
i.e., the vector space representation of language in LLMs, which is a core
element in the deep learning approach to language, shares certain properties with
Frege’s notion of sense. The sense is objective and can be shared; within the
same context we have the same sense. A word can also carry a sense without a
reference, and the sense of individual parts of a sentence determine the sense of
the entire sentence. These points are well reflected in the vector space
representation of language. The LLM representation of a word or sentence may
have sense but not reference, its word-representations determine the sentence
representation, the representations are non-subjective, and in the same context
we have the same word representation. The vector space representation, being
an abstract conceptual representation, also does not operate at the text level, just
as Frege’s sense which is neither on the level of signs, i.e., the text level, nor in
the world of objects and facts; the sense resides at an intermediate level between
the actual signs and the references. Finally, the sense is understood by those with
sufficient language knowledge, i.e., sufficient competence to communicate,
read, write, etc. If this competence is attributed to the LLM, then the LLM
understands the sense of a word or a sentence. In regard to the points 1.-8.
mentioned above, this seems to be the only point which may be contested as we
may deny LLMs linguistic competence. Frege's writings however suggest that
the requirement for linguistic competence is relatively low, as it is distinct from
the notion of knowledge, which involves the concept of truth.

The sense is neither the actual sign nor the representation of that specific sign
on its own. The sense is a representation, a point in a high-dimensional vector
space, but this point only represents in conjunction with other points, within the entire model’s representation of language. Only the joint representation is
meaningful and gives individual representations their sense. In a certain sense,
then, this concept goes beyond Russell and Frege, who relate the sense of a
proper name to the sense of the sentence in which the proper name occurs. In
LLMs, it’s the totality of words and sentences, and their myriad interrelations,
contextualized in many ways, as experienced through the dataset and training,
that essentially gives a word its sense.

Ultimately, sense results from the use of language. Sense can be associated
with the distributed representation of deep learning (DL) models used in natural
language processing (NLP) to handle linguistic tasks. But a single representation
itself is ineffective; as previously stated, it only comes into effect in interaction
with other representations. More so, its effect is realized only in its use or
generation mode. Without appropriate decoding, without a mechanism that
brings the representation to life, the representation itself does not mean much, it
is merely a vector space representation, even though it contains all the
information learned through training. The impact of the representation is
revealed only through its use.

How meaning should be understood has puzzled authors for decades, in
particular in regard to sense and reference. According to our analysis, we have
now a mathematical framework and technical implementation that potentially
represents the sense of a word or sentence. This does not mean that we find
exactly Frege’s sense here; inter alia Frege could not possibly have this in mind.
Yet, according to the above considerations, the latent space representation of
LLMs does share similarities with Frege’s sense; we may therefore hope that
this technical analysis may contribute to elucidating Fregean sense.

When it comes to Frege’s reference we face the same difficulties as discussed
in the context of Russell’s theory. Understanding a sentence means grasping its
sense (thought) and its reference (truth value). We gain knowledge with a
judgment, advancing from thought to truth value. Language models like
ChatGPT operate in a textual world where words and sentences (representations)
can directly refer only to other words and sentences (representations). With
Frege, words can refer to other words too, but only in cases of indirect speech.
In indirect speech, words are used indirectly or have an indirect reference, where
the indirect reference of a word is its usual sense. Thus, LLMs seem to treat
words always as if they have an indirect reference. We may go beyond this
notion of reference, as mentioned above, but doing so requires further study.

Reference is a key aspect of meaning, but it is not the only aspect. Frege
makes a compelling case that we need meaning as sense too. Without it we
would face the logical and epistemological puzzles extensively discussed in the
literature. Russell stresses the contextual nature of meaning; it is the context of
a word that makes it meaning; supplementing this notion with a logical analysis.
With Frege it seems that this goes both ways: words give the sentence its
meaning and the sentence impacts the meaning of words. In any case, the notion
of truth, which pertains to the sentence, is fundamental. We should therefore ask
whether or how LLMs can refer to the truth, rather than how words have
reference of their own, although this may be part of determining the truth. The
classical semantic theories are embedded in an attempt to provide a logical
description of mathematics, language, and perhaps the world, where the concept of truth is central. In ordinary language use, truth plays a much smaller role, as
we often experience. It should be clear that truth is just one of many features that
plays a role in communication. However, it can be the decisive feature, also for
LLMs in critical applications.

\section{Conclusion}
In the “octopus test” Bender and Koller \citep{bender2020} present an
illustration of the limitations of large language models (LLMs) in understanding
word meanings. They describe a scenario where an octopus, having learned
language solely through eavesdropping, lacks true meaning because it cannot
access the referents of the words. The octopus, that is to say LLMs, trained
purely on form, i.e., text, will never learn meaning because they lack reference.

According to our analysis, the question “Do large language models (LLMs)
understand?” is more nuanced; it depends on how the question and the involved
terms are interpreted. The kinds of meaning involved and the internal
representations and workings of LLMs need to be examined. We provide a more
precise take on the question with a dual approach: concisely describing how
LLMs function and generate representations and narrowing down the question
in regard to their semantics such that we can take recourse to well-established
theories of meaning.

Although the development of AI and LLMs progresses rapidly, it is indeed
possible to identify stable elements in the underlying methodology. LLMs are
based on a probabilistic and connectionist approach that generates a distributed
representation of language, which, when scaled up, shows emergent downstream
capabilities.

Whether these LLMs exhibit semantic capabilities, is explored through the
classical semantic theory which goes back to Frege and Russell. We show that
the answer depends on how meaning is defined by Frege and Russell (and which
interpretation one follows). If meaning is solely based on reference, that is, some
referential capability, LLM-generated representations are meaningless, because
the text-based LLMs representation do not directly refer to the world unless the
reference is somehow indirectly induced. Also the notion of reference hinges on
the notion of truth; ultimately it is the truth that determines the reference. If
meaning however is associated with another kind of meaning such as Frege’s
sense in addition to reference, it can be argued that LLM representations can
carry that kind of semantics.

The classical semantic theory, the theory of meaning originating with Frege
and Russell, emphasizes an objective concept of meaning and truth. Terms like
intentions, beliefs, or even consciousness are not directly involved in the
discussion. A corresponding analysis in terms of those concepts may therefore
come to different conclusions. However, we believe that the approach has some
validity, because we would like LLMs to speak validly about reality and to have
truth, even if we and future LLMs cannot say exactly what this consists of.

\section*{Acknowledgements}
I would like to thank Philippe Huneman, Alexandre Gefen and the
participants of the seminars and workshops on Philosophy of AI/Science/LLMs at IHPST Paris, University of Berne and New York University for feedback and
discussion on earlier drafts of this paper.

%

%

%
%
%


\bibliographystyle{unsrtnat}


\end{document}